\documentclass{article}

\usepackage[preprint]{_sty/neurips_2026}

\usepackage[utf8]{inputenc} 
\usepackage[T1]{fontenc}    
\usepackage{hyperref}       
\usepackage{url}            
\usepackage{booktabs}       
\usepackage{amsfonts}       
\usepackage{nicefrac}       
\usepackage{microtype}      
\usepackage{xcolor}         
\usepackage{comment}
\usepackage{graphicx}
\usepackage{amsmath}
\usepackage{bbm}
\usepackage{pifont}
\usepackage{amssymb}
\usepackage{makecell}
\usepackage{wrapfig}
\usepackage{booktabs}

\title{Rollback-Free Stable Brick Structures Generation}

\author{%
  Chenhui Xu$^{1,2}$ \quad Ziyue Bai$^{1,2}$ \quad Fuxun Yu$^3$ \quad Heng Huang$^2$ \quad Jinjun Xiong$^{1,4}$\\
  $^1$ University of Buffalo \quad $^2$ University of Maryland, College Park \\
  $^3$ Microsoft \quad $^4$ University of Texas at San Antonio\\
  \texttt{\{cxu26,jinjun\}@buffalo.edu} \\
}

\begin{document}

\maketitle

\begin{abstract}

While autoregressive models have advanced 3D generation, creating physically stable brick structures remains a challenge due to the strict requirements of gravity and interconnectivity. Existing approaches rely on external physical simulators during inference to perform rejection sampling and brick-by-brick rollbacks, which severely bottlenecks efficiency. To address this, we propose a reinforcement learning paradigm that shifts physical validity enforcement from test-time correction to training-time policy optimization. By utilizing assembly-level rewards, the model optimizes for collision avoidance, global connectivity, structural interlocking, and shape conformity. This paradigm allows the model to internalize physical priors, enabling the first rollback-free generation of stable brick structures. Experimental results demonstrate that our approach achieves state-of-the-art generation quality while accelerating inference speed by orders of magnitude. Our code and dataset are available at \url{https://github.com/miniHuiHui/STABLE}. Our models are available at \url{https://huggingface.co/miniHui/STABLE}.
\end{abstract}
\section{Introduction}
While large autoregressive models have driven profound advancements across various generative tasks~\cite{achiam2023gpt, tian2024visual}, their extension to 3D generation~\cite{pmlr-v119-nash20a, li2024general} often overlooks a crucial practical requirement: physical constructibility. Generating structures from standard, discrete components, such as LEGO{\textregistered} bricks, demands strict adherence to physical constrains during the assembly process. To bridge this gap, pioneer works like BrickGPT~\cite{pun2025generating} have explored 3D brick structure generation~\cite{ge2024learn,chung2021brick}.

A critical challenge in this domain is ensuring the physical stability of the generated structures. This requires the model to be capable of perceiving discrete, constrained, and interactive 3D geometry. Unlike virtual meshes or point clouds, physical assemblies must strictly respect gravity, balance, and precise interconnectivity, where a single misplaced brick can lead to structural collapse. 

To address this, existing autoregressive approaches~\cite{pun2025generating} achieve stability by tightly coupling the generation process with external physical simulators during inference time. As shown in Fig.~\ref{fig:intro}, whenever an unstable or colliding placement is detected, the system triggers a brick-by-brick rejection sampling and physics-aware rollback mechanism, forcing the model to discard invalid steps and repeatedly re-sample tokens until a stable configuration is found. Unfortunately, this trial-and-error paradigm is highly counter-intuitive to the nature of autoregressive decoding, since it breaks the continuous sequence generation assumption, leading to the degradation of generation quality or collapse of generative model. Meanwhile, the constant need to halt generation, query a simulator, and discard previously generated tokens severely bottlenecks inference efficiency.

The reliance on inference-time simulator-based correction reflects a fundamental limitation of standard autoregressive training for physically constrained generation. Although the demonstrations in the dataset are physically valid, such principle is only implicitly embedded in the completed structures rather than exposed as explicit supervision. 
Standard autoregressive modeling learns through by-token imitation, encouraging the model to reproduce local sequence patterns and shape-oriented brick placements. As a result, supervised next-token imitation alone provides insufficient pressure for the model to uncover and internalize the hidden construction principles behind stable examples. Motivated by this inherent inefficiency, we raise a fundamental question:


\begin{figure}[t]
  \centering
   \vspace{-1mm}
\includegraphics{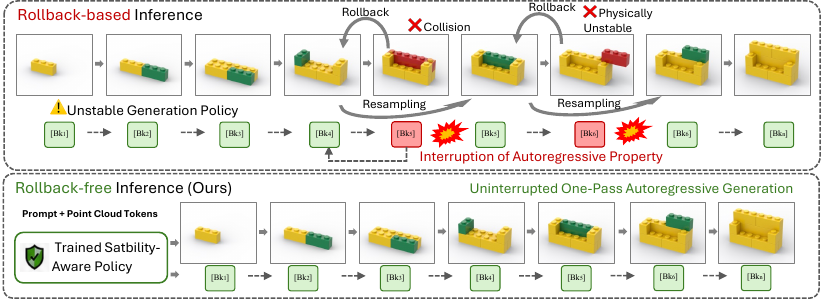}
 \vspace{-4mm}
  \caption{\textbf{Comparison between rollback-based and rollback-free inference}. Rollback-based generation requires resampling after unstable steps, interrupting the autoregressive process, while our stability-aware policy enables uninterrupted one-pass generation.}
  \label{fig:intro}

  \vspace{-6mm}
\end{figure}

\begin{center}
    \textit{Is structural stability inherently learnable by the autoregressive generative model itself?}

\end{center}

In this work, we demonstrate that structural stability can be learned through \textbf{Reinforcement Learning from Physical Rewards (RLPR)}, a novel training paradigm which we propose that optimizes generative models using rewards grounded in physical and construction constraints.
Our key insight is to reformulate stable brick generation from by-token imitation into structure-level policy learning. A generated brick sequence can be viewed as an action trajectory and its final assembly can be evaluated by construction-level objectives through reinforcement learning.
In this way, global feedback from the completed structure can be propagated back to the token decisions that produced it. 

We further decompose assembly-oriented structural stability into several verifiable construction properties. A buildable brick assembly should first satisfy geometric validity, where bricks occupy legal positions without spatial overlap. It should also form a globally coherent structure rather than disconnected or floating fragments. Beyond global coherence, it should exhibit local mechanical robustness, where upper-layer bricks interlock with lower-layer bricks instead of forming weak vertical stacks. These properties together define the stability requirements of brick construction. 

To this end, we propose \texttt{STABLE}, a reinforcement-learning framework that shifts construction validity enforcement from inference-time correction to training-time policy optimization. We instantiate the above decomposition with deterministic assembly-level rewards, including a collision penalty for geometric validity, a connectivity reward for global coherence, an interlocking reward for local robustness, and a shape reward for aligning the target 3D geometry. 
By jointly optimizing these objectives, \texttt{STABLE} turns latent construction principles into explicit learning signals. 

At inference time, \texttt{STABLE} directly emits a complete brick sequence in one uninterrupted autoregressive pass, without rejection sampling, brick-by-brick rollback, or external physical simulation.
Our experiments show that \texttt{STABLE} achieves extremely low levels of collisions and instability while ensuring that the constructed brick assembly closely matches the target. Compared to rollback-based method like BrickGPT~\cite{pun2025generating}, \texttt{STABLE} reduces inference latency by 94\%.

\textbf{Contributions.} In summary, our main contributions are three-fold:
\begin{itemize}
\vspace{-2mm}

\item We challenge the conventional view that physically stable discrete 3D generation must rely on external test-time simulators and greedy rollback correction. Instead, we show that stability-aware and collision-free generation can be learned as an intrinsic capability of the generative policy through Reinforcement Learning from Physcial Rewards (RLPR).
\item We introduce a comprehensive physics- and geometry-aware reward system (incorporating collision, shape, interlocking, and connectivity constraints) that successfully embeds real-world brick structure construction priors into the autoregressive generative model.
\item We demonstrate the first rollback-free framework that achieves structurally stable brick assembly generation without rejection sampling or brick-by-brick rollback, substantially improving inference efficiency while maintaining high geometric and structural quality.
\end{itemize}


\section{Related Works}

\paragraph{Autoregressive Generation for 3D Structures.}
Autoregressive models have become a dominant paradigm for language and visual generation by decomposing complex outputs into sequential token prediction~\cite{achiam2023gpt,tian2024visual}. This idea has also been extended to 3D generation, including mesh modeling in PolyGen~\cite{pmlr-v119-nash20a}, autoregressive point or shape generation~\cite{cheng2022autoregressive,luo2023learning}, and general point model pretraining~\cite{li2024general}. More recent works further explore language-compatible 3D tokenization, allowing continuous shapes to be represented as discrete sequences~\cite{yin2023shapegpt}. However, most of these methods primarily optimize virtual geometric fidelity, where invalid local predictions may harm visual quality but do not necessarily violate physical assembly rules. In contrast, brick-structure generation requires every token to correspond to a valid component placement under strict constraints. 

\paragraph{3D Discretization, Stylization, and Brick Assembly.}
A broad line of 3D learning work studies how to convert continuous geometry into learnable representations. Volumetric methods represent shapes as voxel grids~\cite{wu20153d,maturana2015voxnet}, while sparse and hierarchical structures improve scalability for high-resolution 3D data~\cite{riegler2017octnet}. Point-based models such as PointNet avoid dense voxelization and directly operate on unordered point sets~\cite{qi2017pointnet}. Another related direction is 3D stylization, where differentiable rendering or neural optimization transfers geometric or texture styles across 3D objects and scenes~\cite{kato2018neural,yin20213dstylenet,hollein2022stylemesh}. These works provide useful tools for representing 3D geometry, but they do not directly address the discrete library constraints and physical feasibility requirements of brick-based construction.

Brick assembly introduces a more constrained combinatorial generation problem. 
Early optimization-based methods such as Legolization~\cite{luo2015legolization} generate LEGO brick layouts from input 3D models by jointly considering shape, color, workload, and physical stability through force-based stability analysis and iterative layout refinement. 
Later learning-based works formulate LEGO-like construction as reinforcement learning over discrete components~\cite{chung2021brick}, or study generative pipelines for coherent LEGO micro buildings~\cite{ge2024learn}. 
Most closely related to our work, BrickGPT generates physically stable and buildable brick structures with an autoregressive language model~\cite{pun2025generating}. 
However, these methods either rely on optimization/refinement procedures or use test-time validity checking, rejection sampling, and physics-aware rollback to correct unstable or colliding placements. 
In contrast, we shift physical feasibility from test-time correction to training-enabled policy, enabling rollback-free generation. 


\paragraph{Reward Optimization for Physical Constraint Learning.}
Reinforcement learning and preference optimization are widely used to optimize objectives that are difficult to capture with supervised likelihood alone. PPO stabilizes policy optimization through clipped surrogate objectives~\cite{schulman2017proximal}, while RLHF aligns language models with human preferences after supervised fine-tuning~\cite{ouyang2022training}. GRPO further provides a group-relative optimization strategy that removes the need for a separate value model and has been used for reasoning-oriented model training~\cite{shao2024deepseekmath}. Our method can be viewed as an instance of Reinforcement Learning from Physical Rewards (RLPR).
Unlike RLHF, which relies on human preference feedback, or RLVR, which uses symbolic or
verifiable correctness signals, RLPR optimizes generative models with rewards grounded in physical
or construction constraints.
These rewards propagate delayed physical consequences back to token-level brick decisions, allowing
the model to internalize stability as a generative property.
\section{Dataset}
\label{sec:dataset}

We introduce \textbf{PointCloud2Brick}, a point-cloud-conditioned brick-structure corpus generation dataset constructed from StableText2Brick~\cite{pun2025generating}. 
PointCloud2Brick reformulates stable toy-brick structures as paired \emph{point-cloud-token inputs} and \emph{brick-token outputs}. 
Instead of using natural-language captions as the primary condition, each sample provides a voxelized 3D point cloud that specifies the target geometry, and the model is trained to autoregressively generate a sequence of physical brick units. 
This formulation directly supports geometry-grounded brick generation and provides a unified data interface for both supervised shape reconstruction and reinforcement learning with stability rewards.

\begin{figure}[t]
  \centering
\includegraphics{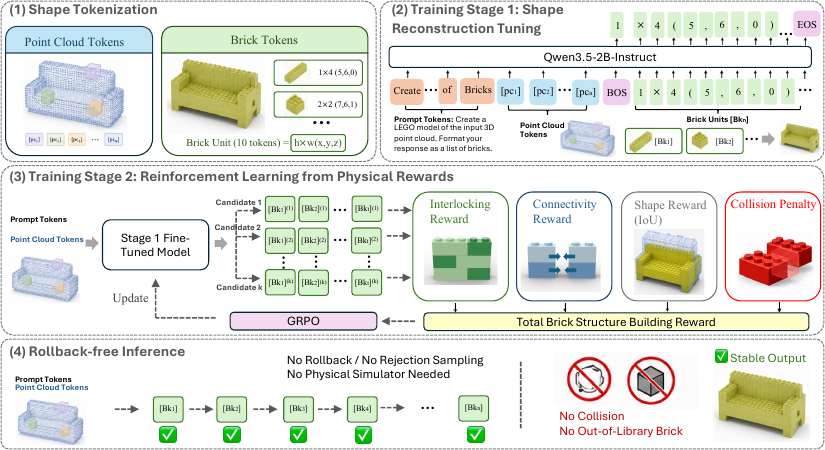}
\vspace{-4mm}
  \caption{\textbf{Overview of \texttt{STABLE}}. \texttt{STABLE} is an autoregressive framework that tokenizes point clouds into LEGO bricks, learns shape reconstruction via SFT, and integrates physical stability into reward signals for reinforcement learning training, and leads to rollback-free inference.}
  \label{fig:main}
  \vspace{-5mm}
\end{figure}

\subsection{PointCloud2Brick Overview}

PointCloud2Brick is constructed from StableText2Brick~\cite{pun2025generating}, which contains 47,389 physically stable brick structures corresponding to 3D objects from 21 ShapeNet~\cite{chang2015shapenet} categories.
Each structure is placed in a discrete $20\times20\times20$ voxel world, where each voxel corresponds to a unit cube. 
All bricks are 1 unit tall and are selected from a library of 8 brick types, yielding 14 oriented variants.


We use the original train/test split and convert each brick structure into a paired point-cloud and brick-sequence representation of~\cite{pun2025generating}. 
The resulting PointCloud2Brick dataset contains 42,604 training samples and 4,785 test samples. 
PointCloud2Brick is designed to support both supervised fine-tuning and reinforcement learning, where the latter additionally uses voxel targets for rewards.

\subsection{Point Cloud Tokenization}
\label{sec:dataset:pc_tokenization}

The input side of PointCloud2Brick is a tokenized point cloud derived from the voxel occupancy of the target structure. 
Given a ground-truth brick layout, we first rasterize all bricks into a binary occupancy grid
$\mathbf{V}\in\{0,1\}^{20\times20\times20}$.
We then enumerate all occupied voxel coordinates:
\begin{equation}
    \mathcal{P}=\{(x_i,y_i,z_i)\mid \mathbf{V}[x_i,y_i,z_i]=1\}.
\end{equation}
Each occupied coordinate is serialized as a point token in the form \texttt{(x,y,z)}, and the full point cloud is represented as a sequence of coordinate tokens:
\(
    \texttt{(x}_1\texttt{,y}_1\texttt{,z}_1\texttt{), (x}_2\texttt{,y}_2\texttt{,z}_2\texttt{), \ldots, (x}_n\texttt{,y}_n\texttt{,z}_n\texttt{)}.
\)

This tokenization preserves the target geometry at voxel resolution while avoiding any assumption about the underlying brick decomposition. 
The conversion from a brick structure to a point cloud is deterministic, but the inverse problem is not unique: multiple valid brick assemblies may correspond to the same occupied voxel set. 
Therefore, the model must do more than copy coordinates; it must infer a compact, legal, and physically plausible decomposition into available brick units.



\subsection{Brick Tokenization}
\label{sec:dataset:brick_tokenization}

Follow~\cite{pun2025generating}, the output side of PointCloud2Brick is an autoregressive sequence of brick tokens. 
Each brick is represented by its oriented footprint and its integer anchor position in the voxel world:
\begin{equation}
    Bk_t = (d_t, p_t), \qquad d_t \in \mathcal{D}, \quad p_t=(x_t,y_t,z_t),
\end{equation}
where $\mathcal{D}$ denotes the allowed set of oriented brick dimensions and $p_t$ specifies the brick position as described in Appendix~\ref{apx:data}. 
In text form, each brick unit is serialized as:
\(
\texttt{h$\times$w (x,y,z)}.
\)
For example, as shown in Fig.~\ref{fig:main}, a brick with footprint $1{\times}4$ placed at $(5,6,0)$ is written as:
\(
\texttt{1x4 (5,6,0)}.
\)

A complete brick structure is therefore represented as:
\(
\mathbf{B} = [Bk_1,Bk_2,\ldots,Bk_T],
\)
where each generation step predicts one brick token conditioned on the point cloud tokens and all previously generated brick tokens. 
This representation naturally matches the sequential construction process: the model does not output a dense voxel grid, but instead generates an ordered list of brick components.







\section{STABLE Framework}
\label{sec:method}

The core idea of \texttt{STABLE} is to convert physical feasibility of autoregressive brick generation from a test-time correction problem into a training-time policy learning problem. To accomplish this goal, we design a two-stage post-training framework as shown in Fig.~\ref{fig:main}. The first stage provides the model with a reliable brick construction prior, while the second stage focuses on improving the physical stability of the generated brick assembly. This design is motivated by a key limitation of supervised brick generation. A model trained only to imitate ground-truth brick sequences can learn the output format and approximate the target geometry, but it is not explicitly optimized for the consequences of its own free-running generations. Small local errors may accumulate into voxel collisions, disconnected components, or weakly interlocked assemblies. \textsc{\texttt{STABLE}} addresses this gap by combining shape-oriented supervised tuning with stability-oriented reinforcement learning. 



\subsection{Stage I: Shape Reconstruction with Supervised Fine-Tuning}
\label{sec:method:sft}

The first stage aims to scaffold the model with the basic capability of translating a target point cloud into a valid brick sequence.
We fine-tune a pretrained instruction-following language model (\texttt{Qwen3.5-2B}~\cite{yang2025qwen3} in our experiments) on PointCloud2Brick, where each sample contains (1) a structured instruction prompt, (2) a point-cloud prompt and (3) a target brick layout written in the prescribed textual format. This stage is not intended to solve physical stability by itself. 

Instead, it teaches the model the grammar of brick generation, the use of the allowed brick library, and the coarse correspondence between voxelized shapes and brick decompositions.
The supervised model provides a stable starting distribution from which the model can generate parseable and shape-relevant candidates, which serves as an initialization for later policy learning. This is crucial for reinforcement learning: if the initial policy cannot produce meaningful brick structures, the reward signals become sparse and noisy. We will show its indispensability later in experiments in Section~\ref{sec:experiments}. 

\subsection{Stage II: Reinforcement Learning from Physical Rewards}
\label{sec:method:rl}

Although the brick sequences data used in supervised fine-tuning are derived from physically stable structures, their stability information is only implicitly and indirectly encoded in the autoregressive ordering of bricks. SFT only trains the model to imitate the reference autoregressive sequence. It does not evaluate the model's own free-running generations, nor can it propagate the physical consequences of a completed structure back to the token decisions that produced it. Therefore, even if the training targets are stable, the model may still generate sequences whose final assemblies contain collisions, disconnected components, or weak inter-layer support. The key limitation is not the absence of stability in the data, but the absence of an explicit mechanism that tells the model how its generated bricks affect the physical validity of the whole structure.

To address this limitation, our \texttt{STABLE} introduces RLPR, a reinforcement-learning-based approach that uses the model's own generated assemblies as the object of optimization. As shown in Fig~\ref{fig:main}, we first decompose physical stability into several verifiable structural properties, such as spatial feasibility (collision), interlocking, connectivity, and shape consistency. 
These properties serve as the feedback interface for reinforcement learning, while their concrete reward definitions are detailed in Section~\ref{sec:reward}. For each point-cloud prompt, the policy samples multiple candidate brick sequences; each completed sequence is parsed into a brick assembly and assigned structure-level feedback. We then use Group Relative Policy Optimization (GRPO)~\cite{guo2025deepseek} to propagate this feedback along the sampled autoregressive trajectory, increasing the likelihood of token sequences that lead to more stable structures and suppressing those that lead to invalid ones. In this way, delayed physical consequences are transformed into token-level policy updates that govern the entire generation process.

\subsection{Rollback-Free Decoding}
\label{sec:method:inference}

After the two-stage training process, inference follows ordinary autoregressive decoding. Given a point-cloud prompt, the model directly emits a complete brick sequence without simulator calls, rejection sampling, or brick-by-brick rollback. This preserves the causal continuity of autoregressive generation and avoids the efficiency bottleneck of test-time correction. The generated structure is therefore the direct output of a stability-aware policy learned during training.

\begin{figure}
    \centering
    \includegraphics{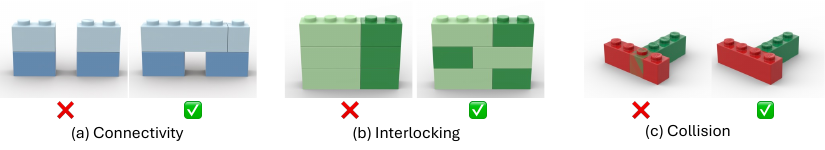}
    \vspace{-7mm}
    \caption{Illustration of rewards related to physical stability}
    \label{fig:reward}
    \vspace{-4mm}
\end{figure}

\section{Reward Design}
\label{sec:reward}

As the core part of reinforcement learning, the reward module is designed to translate construction requirements into scalar learning signals for policy optimization. Given a generated text sequence $\mathbf{y}$ conditioned on input cloud point tokens $\mathbf{x}$, we first parse it into a brick structure $\hat{\mathcal{B}}$. Invalid outputs, including empty responses and malformed brick specifications, are treated as failed constructions. For valid structures, we rasterize all generated bricks into an occupancy grid. Let $C_{\hat{\mathcal{B}}}(v)$ denote the number of generated bricks occupying voxel $v$, and let
\begin{equation}
    \hat{V}(v) = \mathbbm{1}[C_{\hat{\mathcal{B}}}(v) > 0]
\end{equation}
be the binary occupancy grid of the generated structure. The target voxel grid is denoted as $V^{*}$. The total reward is the sum of four terms, which will be ranging from $[-10, 10]$:
\begin{equation}
    R(\mathbf{y}, \mathbf{x})
    =
    R_{\mathrm{col}}(\mathbf{y})
    +
    R_{\mathrm{shape}}(\mathbf{y}, \mathbf{x})
    +
    R_{\mathrm{inter}}(\mathbf{y})
    +
    R_{\mathrm{conn}}(\mathbf{y}).
    \label{eq:total_reward}
\end{equation}

\textbf{Collision Penalty.}
The collision term penalizes voxel-level overlaps between bricks. A collision occurs when more than one brick occupies the same voxel, as shown in Fig~\ref{fig:reward}(c). Let
\begin{equation}
    N_{\mathrm{col}}
    =
    \sum_{v}
    \mathbbm{1}
    [C_{\hat{\mathcal{B}}}(v) > 1]
\end{equation}
be the number of colliding voxels. The collision reward is defined as
\begin{equation}
    R_{\mathrm{col}}(\mathbf{y})
    =
    \max(-10, -2N_{\mathrm{col}})\in [-10,0]
 \label{eq:collision_reward}
\end{equation}
This term acts as a hard feasibility pressure. Even a small number of collisions receives a direct penalty, while severe collision failures are clipped to avoid excessively large negative rewards for the stabilization of RL training. Because brick assemblies require exact discrete placement, this penalty is essential for preventing the model from exploiting shape overlap at the cost of physical invalidity.

\textbf{Shape Conformity Reward.}
\label{sec:reward_shape}
The shape reward measures how well the generated brick structure approximates the target point-cloud geometry after voxelization. The primary purpose of the shape reward is to guarantee that the model's ability to generate shapes of interest is preserved while optimizing other physics-related objectives. We use volumetric intersection-over-union:
\begin{equation}
    \mathrm{IoU}(\hat{V}, V^{*})
    =
    \frac{
    \sum_v \mathbbm{1}[\hat{V}(v)=1 \wedge V^{*}(v)=1]
    }{
    \sum_v \mathbbm{1}[\hat{V}(v)=1 \vee V^{*}(v)=1]
    }.
\end{equation}
The reward is scaled as
    \vspace{-4mm}
\begin{equation}
    R_{\mathrm{shape}}(\mathbf{y}, \mathbf{x})
    =
    5 \cdot \mathrm{IoU}(\hat{V}, V^{*})\in [0,5]
    \label{eq:shape_reward}
\end{equation}
Unlike the structural rewards below, the shape reward is not gated by collision-freeness. This choice provides a stable geometric learning signal throughout training, even when the early policy still produces imperfect structures. The collision penalty then counterbalances degenerate solutions that achieve high occupancy overlap by placing mutually intersecting bricks.

\textbf{Interlocking Reward.}
\label{sec:reward_interlocking}
Collision-free shape matching does not necessarily imply mechanical robustness. A structure may occupy the correct voxels while still forming simple vertical stacks, where each upper brick is supported by only one brick below. Such layouts are geometrically valid but mechanically weak, since bricks are not tied across seams. As shown in Fig~\ref{fig:reward}(b), the interlocking reward therefore encourages upper-layer bricks to bridge multiple lower-layer bricks.
For each generated brick $b_i$, let $z_i$ denote its layer index and let $\Phi(b_i) \subset \mathbb{Z}^2$ denote its horizontal footprint:
\begin{equation}
    \Phi(b_i)
    =
    \{(x,y) \mid x_i \leq x < x_i + h_i,\;
    y_i \leq y < y_i + w_i \}.
    \label{eq:brick_footprint}
\end{equation}
We define the set of lower-layer supporting bricks for $b_i$ as
\begin{equation}
    \mathcal{S}^{-}(b_i)
    =
    \left\{
    b_j \;\middle|\;
    z_j = z_i - 1,\;
    \Phi(b_i) \cap \Phi(b_j) \neq \emptyset
    \right\}.
    \label{eq:lower_support_set}
\end{equation}
This mapping converts the generated brick layout into a cross-layer support relation: an upper brick is considered supported by a lower brick if their horizontal footprints overlap and the lower brick lies exactly one layer below.
A brick is considered interlocked if it bridges at least two distinct supporting bricks in the layer below:
    \vspace{-4mm}
\begin{equation}
    I_{\mathrm{inter}}(b_i)
    =
    \mathbf{1}
    \left[
    z_i > 0
    \;\wedge\;
    |\mathcal{S}^{-}(b_i)| \geq 2
    \right].
    \label{eq:interlocking_indicator}
\end{equation}
This criterion captures the common brick-building principle that an upper brick should cover seams between lower bricks rather than simply stack on top of a single brick. Ground-layer bricks are excluded because they do not have lower-layer supports.
The structure-level interlocking score is the fraction of non-ground-layer bricks that satisfy this bridging condition:
\begin{equation}
    S_{\mathrm{inter}}(\hat{\mathcal{B}})
    =
    \frac{
    \sum_{b_i \in \hat{\mathcal{B}}}
    I_{\mathrm{inter}}(b_i)
    }{
    \max
    \left(
    \sum_{b_i \in \hat{\mathcal{B}}}
    \mathbf{1}[z_i > 0],
    1
    \right)
    }.
    \label{eq:interlocking_score}
\end{equation}

\vspace{-3mm}
The interlocking reward is then
\(R_{\mathrm{inter}}(\mathbf{y})=
    3 \cdot S_{\mathrm{inter}}(\hat{\mathcal{B}})\in [0,3]\), if \(\hat{\mathcal{B}}\) is collision-free and in bounds.


\textbf{Connectivity Reward.}
\label{sec:reward_connectivity}
The connectivity reward encourages the generated structure to form a single coherent assembly rather than a set of disconnected or floating components, as shown in Fig~\ref{fig:reward}(a). We compute connectivity over the occupied voxels or their corresponding brick-support graph. Let $\mathcal{O}$ be the set of occupied voxels in the generated structure and let $\mathcal{D}$ be the subset of occupied voxels that are disconnected from the main grounded component. The connectivity score is
\begin{equation}
    S_{\mathrm{conn}}(\hat{\mathcal{B}})
    =
    1 -
    \frac{
    |\mathcal{D}|
    }{
    \max(|\mathcal{O}|, 1)
    }.
\end{equation}
\vspace{-3mm}

The corresponding reward is
\(
    R_{\mathrm{conn}}(\mathbf{y})
    =
    2 \cdot S_{\mathrm{conn}}(\hat{\mathcal{B}})\in [0,2]\),
   if \(\hat{\mathcal{B}}\) is collision-free and in bounds.

       \vspace{-3mm}



\section{Experiments}
\label{sec:experiments}
    \vspace{-2mm}
\textbf{Setup.} We implement \texttt{STABLE} as a two-stage post-training pipeline on top of \texttt{Qwen3.5-2B}~\cite{yang2025qwen3}.
We first conduct supervised fine-tuning and obtain an SFT intermediate model \texttt{STABLE-SFT}. Starting from \texttt{STABLE-SFT} model, we perform GRPO training and conclude the final \texttt{STABLE} model. 
We use full-parameter fine-tuning for both SFT and GRPO. The model is trained on 4 NVIDIA H100/H200 GPUs with vLLM~\cite{kwon2023efficient} acceleration for on-policy generation and DeepSpeed~\cite{rasley2020deepspeed} ZeRO-2 for memory-efficient optimization. We conduct full-
parameter fine-tuning on the model for more stable conver-
gence and higher final accuracy. Training details are provided in the Appendix~\ref{app:training-details}.




\textbf{Evaluation.} We evaluate all models on samples from the PointCloud2Brick test split. 
All methods are decoded with greedy decoding and a maximum generation length of 8,192 tokens. 
We evaluate each generated sequence at two levels. 
At the geometric level, we report voxel IoU between the generated occupancy and target occupancy grid. At the physical-structure level, we report collision-free rate, connectivity ratio, interlocking score, and seam coverage rate. See details in Apendix~\ref{app:evaluation-details}.


    \vspace{-2mm}
\subsection{Stable Brick Construction Results}
\label{sec:exp:main}
    \vspace{-1mm}
\textbf{Baselines.} We use \texttt{Gemini-3.1-Pro}~\cite{team2023gemini} and \texttt{Llama-4-Scout-17B-16E-Instruct}~\cite{touvron2023llama} as representative close- and open-source LLM baseline separately. We compare our model with other brick generation models like BrickGPT~\cite{pun2025generating} and LegoACE~\cite{xu2025legoace}. All the models are under the same rollback-free decoding protocol, except BrickGPT+RB\&RS (rollback and rejection sampling). We also evaluate on \texttt{Qwen3.5-122B-A10B}~\cite{yang2025qwen3} to show the impact of model scales. 

\textbf{Remark 1: General-purpose LLMs fail to directly solve stable brick construction.}
As shown in Table~\ref{tab:main_results}, prompting general LLMs such as \texttt{Llama-4}, \texttt{Qwen3.5-122B-A10B}, and \texttt{Qwen3.5-2B} results in nearly zero valid brick generation.
Even stronger instruction-following models like \texttt{Gemini-3.1-Pro} do not naturally understand the discrete brick grammar, physical constraints, or point-cloud-to-brick decomposition required by this task. This indicates that generating brick assemblies is fundamentally challenging for LLMs regardless of model capacity and scale.

\begin{figure}[t]
    \centering
    \includegraphics[width=\linewidth]{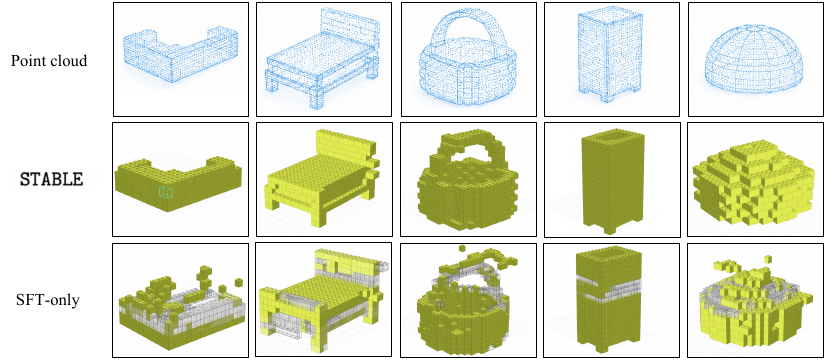}
        \vspace{-7mm}
    \caption{Qualitative results of Point Cloud to Brick generations. Transparency indicates a collision.}
    \vspace{-6mm}
    \label{fig:results}
\end{figure}

\textbf{Remark 2: SFT learns shape, but not physical feasibility.}
As shown in Table~\ref{tab:main_results}, \texttt{STABLE-SFT} achieves a high voxel IoU of 0.813, showing that supervised fine-tuning teaches the model to reconstruct target geometry.
However, its collision-free rate remains 0.00, and its structural metrics are almost zero.
The qualitative examples in Fig.~\ref{fig:results} show the same pattern: \texttt{STABLE-SFT} outputs roughly cover the target shape but contain redundant, overlapping, and weakly connected bricks.
This confirms that maximum-likelihood imitation alone is insufficient for stable construction. At the same time, \texttt{STABLE-ZERO}, trained directly from the base model, still scores zero on all metrics, meaning SFT remains an essential component since RL alone can not tell the model how to build bricks.

\textbf{Remark 3: RL is the key to learning physical stability.}
After GRPO training, \texttt{STABLE} improves the collision-free rate to 0.99 while maintaining strong voxel IoU of 0.907.
It also achieves the strongest rollback-free structural metrics, including 1.000 mean stability, 1.000 min stability, 0.97 connectivity ratio, 0.722 interlocking score, and 0.676 seam coverage. These results show that our RL training is the decisive step for stable generation, turning a model that merely reconstructs shapes into one that actively prefers collision-free, connected, and interlocked brick assemblies.

\textbf{Remark 4: STABLE outperform rollback-based BrickGPT with much lower inference cost.} As shown in Table~\ref{tab:main_results},
with external rollback and rejection sampling, BrickGPT can reach 1.00 collision-free rate (algorithm enforced) but requires 895.41 seconds per sample due to the calling of external physical simulator.
In contrast, \texttt{STABLE} achieves comparable or better stability, while using only 57.64 seconds per sample.
This demonstrates that moving physical evaluation from inference-time correction to training-time policy learning yields both stronger generation quality and faster inference.

\begin{table*}[t]
\centering
\caption{\textbf{Quantitative results.} Main comparison on PointCloud2Brick test samples. All models are evaluated with greedy decoding and without test-time rollback or rejection sampling except (BrickGPT+RB\&RS). Mean and Min Stability are from simulator-based statistics as in~\cite{pun2025generating}.}
\label{tab:main_results}
\resizebox{\textwidth}{!}{%
\begin{tabular}{lccccccccc}
\toprule
\textbf{Method}

& \makecell{Coll.-Free\\Rate}
& \makecell{Voxel\\IoU}
& \makecell{Mean\\Stability}
& \makecell{Min\\Stability}
& \makecell{Conn.\\Ratio}
& \makecell{Interlock.\\Score}
& \makecell{Seam\\Cov.}
& \makecell{Mean\\Bricks}
& \makecell{Avg. Time\\(s/sample)} \\
\midrule

BrickGPT~\cite{pun2025generating}
 & 0.10 & 0.039 & 0.214 & 0.031 & 0.00 & 0.023 & 0.056 & 110.9 & 32.97 \\
BrickGPT+RB\&RS
 & 1.00 & 0.349 & 0.983 & 0.781 & 0.54 & 0.148 & 0.159 & 168.6 & 895.41 \\
 LegoACE~\cite{xu2025legoace} & 0.05 & 0.080 & 0.041 & 0.007 & 0.12 & 0.103 & 0.086 & 79.3 & -- \\
\midrule
\texttt{Gemini-3.1-Pro}
 & 0.02 & 0.276 & 0.913 & 0.643 & 0.00 & 0.119 & 0.074 & 64.32 & -- \\
\texttt{Llama-4}~\cite{touvron2023llama}
 & 0.00 & 0.000 & -- & -- & 0.00 & 0.000 & 0.000 & 0.0 & 563.12 \\
\texttt{Qwen3.5-122B-A10B}
 & 0.00 & 0.000 & -- & -- & 0.00 & 0.000 & 0.000 & 0.0 & 412.74 \\
\midrule
\texttt{Qwen3.5-2B} (base)
 & 0.00 & 0.000 & -- & -- & 0.00 & 0.000 & 0.000 & 0.0 & 143.89 \\
\texttt{STABLE-ZERO}
 & 0.00 & 0.000 & -- & -- & 0.00 & 0.000 & 0.000 & 0.0 & -- \\
\texttt{STABLE-SFT}
 & 0.00 & 0.813 & 0.947 & 0.228 & 0.05 & 0.012 & 0.010  & 661.5 & 183.63 \\
\texttt{STABLE}
 & \textbf{0.99} & \textbf{0.907} & \textbf{1.000} & \textbf{1.000} & \textbf{0.97} & \textbf{0.722} & \textbf{0.676} & 216.5 & 57.64 \\
\bottomrule
\end{tabular}%
}
    \vspace{-6mm}
\end{table*}

\begin{figure}
    \centering
    \includegraphics[width=\linewidth]{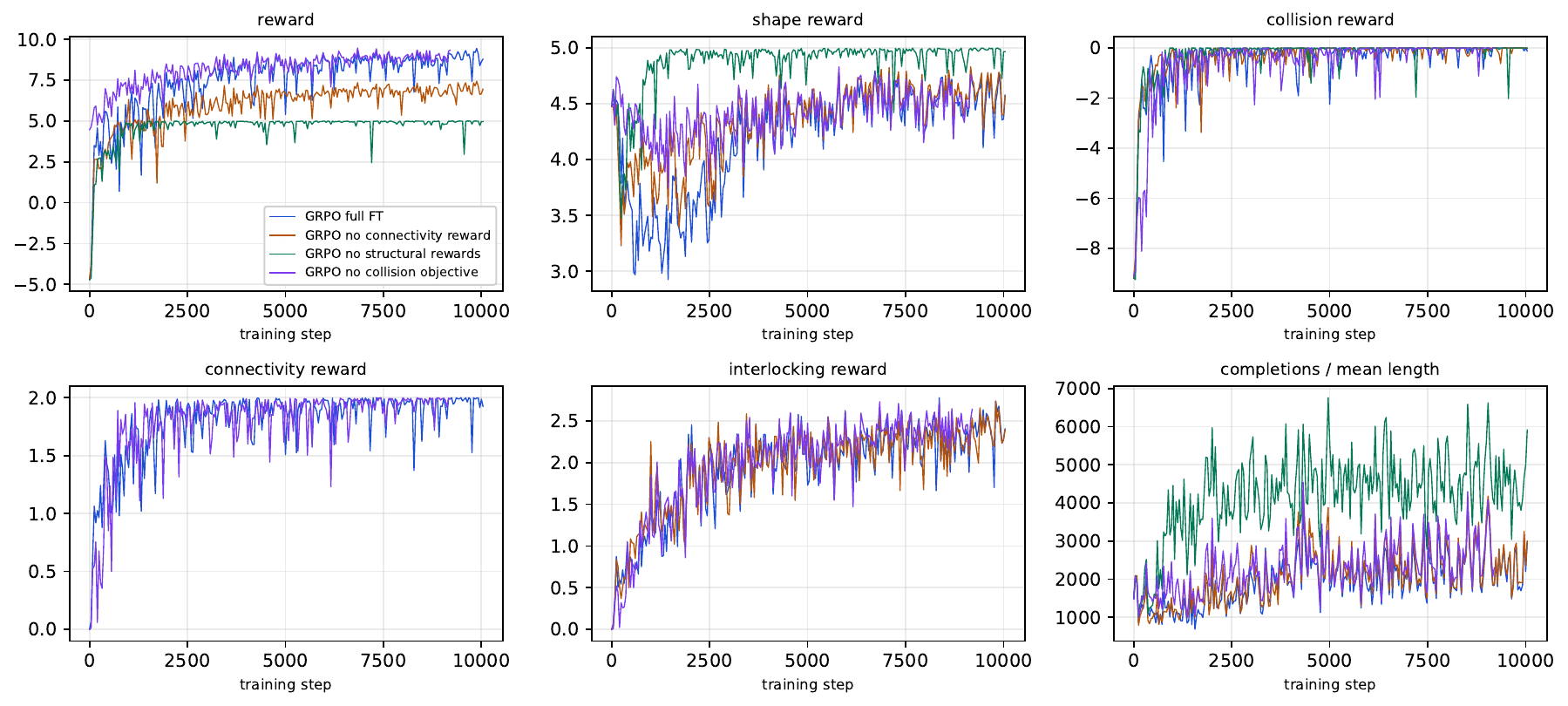}
       \vspace{-7mm}
    \caption{GRPO training dynamics under different reward settings.}
       \vspace{-5mm}
    \label{fig:training_dynamics}
\end{figure}

\begin{table*}[t]
\centering
\caption{Reward ablation on PointCloud2Brick test samples. All variants are initialized from the same SFT checkpoint and trained with GRPO.}
\label{tab:reward_ablation}
\resizebox{\textwidth}{!}{%
\begin{tabular}{lccccccccc}
\toprule
\textbf{Variant}
& \makecell{Coll.-Free\\Rate}
& \makecell{Mean Coll.\\Voxels}
& \makecell{Voxel\\IoU}
& \makecell{Conn.\\Ratio}
& \makecell{Connected\\Rate}
& \makecell{Interlock.\\Score}
& \makecell{Seam\\Cov.}
& \makecell{In-Bounds\\Rate}
& \makecell{Mean\\Bricks} \\
\midrule
\textbf{\texttt{STABLE}}
& \textbf{0.99} & \textbf{0.05} & 0.907 & \textbf{0.97} & \textbf{0.49} & 0.722 & \textbf{0.676} & \textbf{1.00} & 216.5 \\
w/o Collision
& 0.72 & 14.51 & 0.897 & 0.91 & 0.46 & \textbf{0.734} & 0.638 & \textbf{1.00} & 247.1 \\
w/o Connectivity
& 0.93 & 9.50 & 0.919 & 0.73 & 0.45 & 0.709 & 0.653 & \textbf{1.00} & 230.0 \\
w/o Structural
& 0.96 & 12.46 & \textbf{0.978} & 0.46 & 0.02 & 0.069 & \textbf{0.676} & 0.76 & 387.0 \\
\bottomrule
\end{tabular}%
}
   \vspace{-4mm}
\end{table*}
\vspace{-3mm}
\subsection{Training Dynamics}
\vspace{-2mm}
\label{sec:exp:training_dynamics}

Fig.~\ref{fig:training_dynamics} visualizes the RL training dynamics of \texttt{STABLE} and its ablation variants. Details in Appendix~\ref{app:training_dynamics}.

\textbf{Remark 5: RL learns stability without sacrificing shape fidelity.}
As shown in Fig. 5, the total reward increases steadily during GRPO training, while the collision reward quickly approaches zero penalty and the connectivity and interlocking rewards gradually improve. Interestly, the shape reward temporarily drops in the early stage, suggesting that the policy first departs from purely shape-oriented imitation to explore more construction-valid brick decompositions. It then recovers as training progresses, indicating that STABLE learns to satisfy stability constraints while preserving target-shape fidelity. Overall, RL transforms the SFT model from a shape imitator into a policy that generates assemblies that are shape-consistent, collision-free, connected, and interlocked.

\section{Ablation Study}

\label{sec:ablation}

\begin{wraptable}{r}{0.35\textwidth}
\vspace{-15mm}
\centering
\caption{Ablation configurations.}
\label{tab:reward_config}
\scriptsize
\setlength{\tabcolsep}{2.5pt}
\renewcommand{\arraystretch}{0.9}
\begin{tabular}{lcccc}
\toprule
\textbf{Variant}
& $R_{\mathrm{col}}$
& $R_{\mathrm{shape}}$
& $R_{\mathrm{inter}}$
& $R_{\mathrm{conn}}$ \\
\midrule
\textbf{\texttt{STABLE}}
& $\checkmark$ & $\checkmark$ & $\checkmark$ & $\checkmark$ \\
w/o Coll.
& $\times$ & $\checkmark$ & $\checkmark$ & $\checkmark$ \\
w/o Conn.
& $\checkmark$ & $\checkmark$ & $\checkmark$ & $\times$ \\
w/o Struct.
& $\checkmark$ & $\checkmark$ & $\times$ & $\times$ \\
\bottomrule
\end{tabular}
\vspace{-1em}
\end{wraptable}

We ablate the reward components of \texttt{STABLE} to understand how each physical signal contributes to the final construction quality.
All variants start from the same \texttt{STABLE-SFT} checkpoint and are trained with GRPO under identical settings.
The reward configurations are summarized in Table~\ref{tab:reward_config}, and the quantitative results are shown in Table~\ref{tab:reward_ablation}.

\textbf{Remark 6: Collision reward is necessary for exact spatial feasibility.}
Removing $R_{\mathrm{col}}$ significantly increases the mean number of colliding voxels from 0.05 to 14.51 and reduces the collision-free rate from 0.99 to 0.72.
Although the model still obtains reasonable connectivity and interlocking scores, structural rewards alone cannot guarantee voxel-level non-overlap.
This confirms that explicit collision supervision is essential for discrete brick placement.

\textbf{Remark 7: Connectivity reward improves global assembly coherence.}
Without $R_{\mathrm{conn}}$, the connectivity ratio drops from 0.97 to 0.73.
The model can still produce locally interlocked structures, but it becomes less reliable at forming a globally coherent assembly.
Thus, connectivity reward provides a complementary global constraint beyond local brick support.

\textbf{Remark 8: Structural rewards prevent voxel-filling degeneration.}
When both connectivity and interlocking rewards are removed, the model reaches the highest voxel IoU of 0.978 but suffers from poor structural quality, with connectivity ratio dropping to 0.46 and interlocking score to 0.069. We further observe that the model tends to exploit the shape reward by using many small 1$\times$1 bricks to fill target voxels, leading to high bircks usage. Such outputs can match the target shape but lose the coherent and interlocked structure required for buildable assemblies. Therefore, structural rewards are necessary to guide the model toward physically meaningful brick constructions.

\section{Conclusion}
\label{sec:conclusion}

We proposed \texttt{STABLE}, a rollback-free framework for stable brick-structure generation from point clouds. By moving physical evaluation from inference to training, \texttt{STABLE} uses structure-level rewards to teach collision avoidance, connectivity, interlocking, and shape conformity. Experiments show strong geometric fidelity and physical stability with substantially lower inference time, suggesting that discrete 3D physical feasibility can be internalized through reward-based post-training.


{\small

\bibliographystyle{abbrv}
\bibliography{_bib/ref}
}

\newpage
\appendix

\section{Details of Dataset}
\label{apx:data}

\subsection{Brick Library}

All bricks are 1 unit tall and are selected from a library of 8 brick types, yielding 14 oriented variants: $1{\times}1$, $1{\times}2$, $2{\times}1$, $1{\times}4$, $4{\times}1$, $1{\times}6$, $6{\times}1$, $1{\times}8$, $8{\times}1$, $2{\times}2$, $2{\times}4$, $4{\times}2$, $2{\times}6$, and $6{\times}2$.

\subsection{Data Instance}

Each point cloud sequence is inserted into a structured instruction prompt:

\begin{quote}
\small\ttfamily
Create a LEGO model of the input 3D point cloud.\\
Format your response as a list of bricks: <brick dimensions> <brick position>, where the brick position is (x,y,z).\\
Allowed brick dimensions are 2x4, 4x2, 2x6, 6x2, 1x2, 2x1, 1x4, 4x1, 1x6, 6x1, 1x8, 8x1, 1x1, 2x2.\\
All bricks are 1 unit tall.\\[0.5em]
\#\#\# Input Point Cloud:\\
(x$_1$,y$_1$,z$_1$), (x$_2$,y$_2$,z$_2$), \ldots, (x$_n$,y$_n$,z$_n$)
\end{quote}

With estimated output:

\begin{quote}
\small\ttfamily
\#\#\# Bricks:\\
h$_1$$\times$w$_1$(x$_1$,y$_1$,z$_1$), h$_2$$\times$w$_2$(x$_2$,y$_2$,z$_2$), \ldots, h$_T$$\times$w$_T$(x$_T$,y$_T$,z$_T$)
\end{quote}

\subsection{Training Formats}
\label{sec:dataset:format}

\paragraph{SFT format.}
For supervised fine-tuning, each PointCloud2Brick sample is formatted as a multi-turn chat transcript compatible with the Qwen chat template. 
The system message is fixed as \texttt{"You are a helpful assistant."}; the user message contains the structured point-cloud prompt described in Section~\ref{sec:dataset:pc_tokenization}; and the assistant message contains the ground-truth brick sequence, with one brick per line in \texttt{h$\times$w (x,y,z)} format. 
This stage teaches the model to translate voxel-level shape tokens into valid brick-token sequences.

\paragraph{GRPO format.}
For reinforcement learning, the prompt contains only the system and user messages, while the assistant response is generated by the policy. 
Each sample is additionally augmented with a \texttt{target\_voxels} field, which stores the ground-truth binary occupancy grid as a base64-compressed NumPy array. 
During training, this target voxel grid is passed to the reward functions to compute shape conformity and other voxel-level objectives. 
Thus, the same point-cloud and brick-token representation supports both supervised shape reconstruction and reward-based physical-stability optimization.

\subsection{Dataset Statistics}
\label{sec:dataset:stats}

Table~\ref{tab:dataset_stats} summarizes the PointCloud2Brick splits used in our experiments. 
All splits are converted from StableText2Brick using the point-cloud and brick-tokenization pipeline described above.

\begin{table}[h]
\centering
\caption{Statistics of PointCloud2Brick. Each sample contains a point-cloud-token input paired with an autoregressive brick-token output sequence.}
\label{tab:dataset_stats}
\begin{tabular}{lcc}
\toprule
\textbf{Split} & \textbf{Train} & \textbf{Test} \\
\midrule
Total unique samples        & 42{,}604 & 4{,}785 \\
ShapeNet categories         & \multicolumn{2}{c}{21} \\
Unique 3D objects            & \multicolumn{2}{c}{$\sim$28{,}000} \\
Voxel world dimension        & \multicolumn{2}{c}{$20 \times 20 \times 20$} \\
Oriented brick variants      & \multicolumn{2}{c}{14} \\
\bottomrule
\end{tabular}
\end{table}


\section{Training Details}
\label{app:training-details}

\paragraph{SFT configuration.}
Table~\ref{tab:sft-config} lists the supervised fine-tuning configuration. SFT is run as full-model
fine-tuning with the TRL SFT entry point, using the point-cloud chat-format dataset and the
ground-truth brick sequence as the assistant response.

\begin{table}[h]
\centering
\caption{Supervised fine-tuning configuration used to train the point-cloud-conditioned brick generator.}
\small
\renewcommand{\arraystretch}{1.08}
\begin{tabular}{ll}
\hline
\textbf{Hyperparameter} & \textbf{Value} \\
\hline
Base model & \texttt{Qwen/Qwen3.5-2B} \\
Maximum sequence length & 8192 \\
Number of processes & 4 \\
Per-device train batch size & 2 \\
Per-device eval batch size & 2 \\
Gradient accumulation steps & 4 \\
Learning rate & $2\times10^{-4}$ \\
Learning-rate scheduler & cosine \\
Warmup steps & 100 \\
Training epochs & 3 \\
Precision & bfloat16 \\
Dataloader workers & 4 \\
Distributed setting & DDP with unused-parameter detection \\
\hline
\end{tabular}

\label{tab:sft-config}
\end{table}

\paragraph{GRPO configuration.}
Table~\ref{tab:grpo-config} lists the reinforcement-learning configuration used after SFT. GRPO starts
from the SFT checkpoint, samples multiple brick completions for each point-cloud prompt, and updates
the full model using TRL's GRPO trainer with vLLM colocated generation and DeepSpeed ZeRO-2.

\begin{table}[h]
\centering
\caption{GRPO configuration used for physical-stability-oriented reinforcement learning after SFT.}
\small
\renewcommand{\arraystretch}{1.08}
\begin{tabular}{ll}
\hline
\textbf{Hyperparameter} & \textbf{Value} \\
\hline
Initialization & SFT checkpoint \\
Trainer & TRL \texttt{GRPOTrainer} \\
Generation backend & vLLM colocate mode \\
vLLM GPU memory utilization & 0.4 \\
Number of generations per prompt & 8 \\
Maximum completion length & 8192 \\
Sampling temperature & 0.7 \\
Per-device train batch size & 2 \\
Gradient accumulation steps & 4 \\
Training epochs & 1 \\
Learning rate & $5\times10^{-6}$ \\
Learning-rate scheduler & cosine \\
Warmup steps & 100 \\
Maximum gradient norm & 1.0 \\
Precision & bfloat16 \\
Gradient checkpointing & enabled, \texttt{use\_reentrant=False} \\
Distributed optimization & DeepSpeed ZeRO-2 \\
Dataloader workers & 4 \\
Dataloader pin memory & enabled \\
\hline
\end{tabular}

\label{tab:grpo-config}
\end{table}

\section{Evaluation Details}
\label{app:evaluation-details}

All the models are evaluated on a Nvidia L40S GPU.

\paragraph{Evaluation data and decoding.}
We evaluate all systems on the held-out PointCloud2Brick test split for
the reported experiments. Each prompt contains a serialized point cloud in the fixed
$20\times20\times20$ voxel world and requests a list of bricks in the form
\texttt{h}x\texttt{w} \texttt{(x,y,z)}. For causal language models, we apply the Qwen chat template and
decode up to 8192 new tokens. Evaluation uses deterministic decoding for comparability across models.

\paragraph{Parsing and validity.}
Generated text is parsed line-by-line into a \texttt{BrickStructure}. A sample is counted as successfully
parsed only if the output contains at least one valid brick line and the complete set of lines can be
converted into legal brick objects. Collision is computed by rasterizing each brick into the voxel grid
and counting voxels with occupancy greater than one. Out-of-bounds bricks and floating components
are tracked separately so that parse success, geometric validity, and physical plausibility can be
distinguished.

\paragraph{Evaluation metric definitions.}
Let $N$ be the number of evaluated prompts, $\mathcal{P}$ be the subset of samples whose generations
are successfully parsed into non-empty brick structures, $\mathcal{C}\subseteq\mathcal{P}$ be the parsed
samples with no voxel collisions, and $\mathcal{B}\subseteq\mathcal{P}$ be the parsed samples whose
bricks are all inside the $20^3$ world. The metrics used in Tables 1 and 2 are:
\begin{itemize}
    \item \textbf{Coll.-Free Rate}: the fraction of generated structures without voxel collisions,
    reported over all evaluated prompts as $|\mathcal{C}|/N$. A collision occurs when two or more
    bricks occupy the same voxel.
    \item \textbf{Mean Coll. Voxels}: the average number of colliding voxels per parsed generation,
    where a colliding voxel has occupancy greater than one. Lower is better.
    \item \textbf{Voxel IoU}: voxel intersection-over-union between the generated occupancy grid
    $V_{\mathrm{gen}}$ and the target occupancy grid $V_{\mathrm{tar}}$,
    $\mathrm{IoU}=|V_{\mathrm{gen}}\cap V_{\mathrm{tar}}|/|V_{\mathrm{gen}}\cup V_{\mathrm{tar}}|$, averaged over
    parsed generations.
    \item \textbf{Mean Stability}: the average simulator-based per-brick stability score given by Gurobi. For a valid
    structure with brick-level scores $\{s_i\}$, we first compute the structure mean
    $\frac{1}{M}\sum_{i=1}^{M}s_i$, then average across evaluated valid structures. Higher is better.
    \item \textbf{Min Stability}: the weakest-brick stability score $\min_i s_i$ for each valid
    structure, averaged across valid structures. This measures whether any brick is close to failure.
    \item \textbf{Conn. Ratio}: the fraction of occupied voxels belonging to the main connected,
    supported component, computed as $1-|D|/\max(|O|,1)$ where $O$ is the occupied voxel set and
    $D$ is the subset disconnected from the grounded component.
    \item \textbf{Connected Rate}: the fraction of parsed generations classified as connected by the
    connectivity check.
    \item \textbf{Interlock. Score}: the fraction of non-ground-layer bricks that bridge at least two
    distinct lower-layer support bricks. This measures local mechanical interlocking.
    \item \textbf{Seam Cov.}: the seam-coverage score, measuring how often seams between bricks in a
    lower layer are covered by bricks in the layer above. Higher values indicate more staggered,
    buildable layouts.
    \item \textbf{In-Bounds Rate}: the fraction of parsed structures whose bricks all lie inside the
    fixed $20\times20\times20$ voxel world, i.e., $|\mathcal{B}|/|\mathcal{P}|$.
    \item \textbf{Mean Bricks}: the average number of generated bricks per parsed structure.
    \item \textbf{Avg. Time}: average wall-clock inference time per sample in seconds, including any
    decoding-time rollback or rejection-sampling procedure when that method uses one.
\end{itemize}

\begin{figure}
    \centering
    \includegraphics[width=1.0\linewidth]{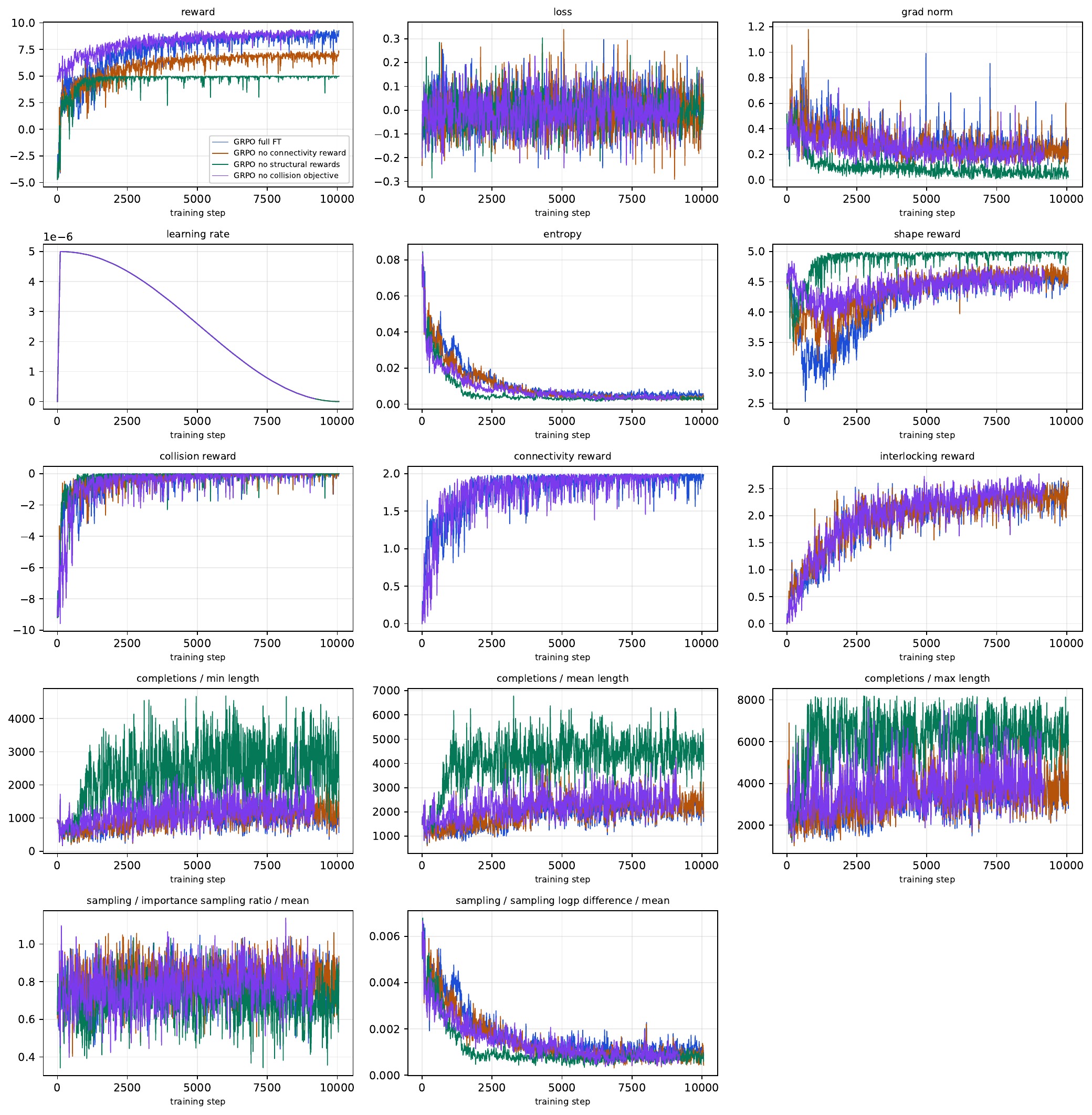}
\caption{Full GRPO training dynamics of \texttt{STABLE} and reward-ablation variants.}
    \label{fig:dynamcis_all}
\end{figure}

\begin{figure}
    \centering
    \includegraphics[width=1.0\linewidth]{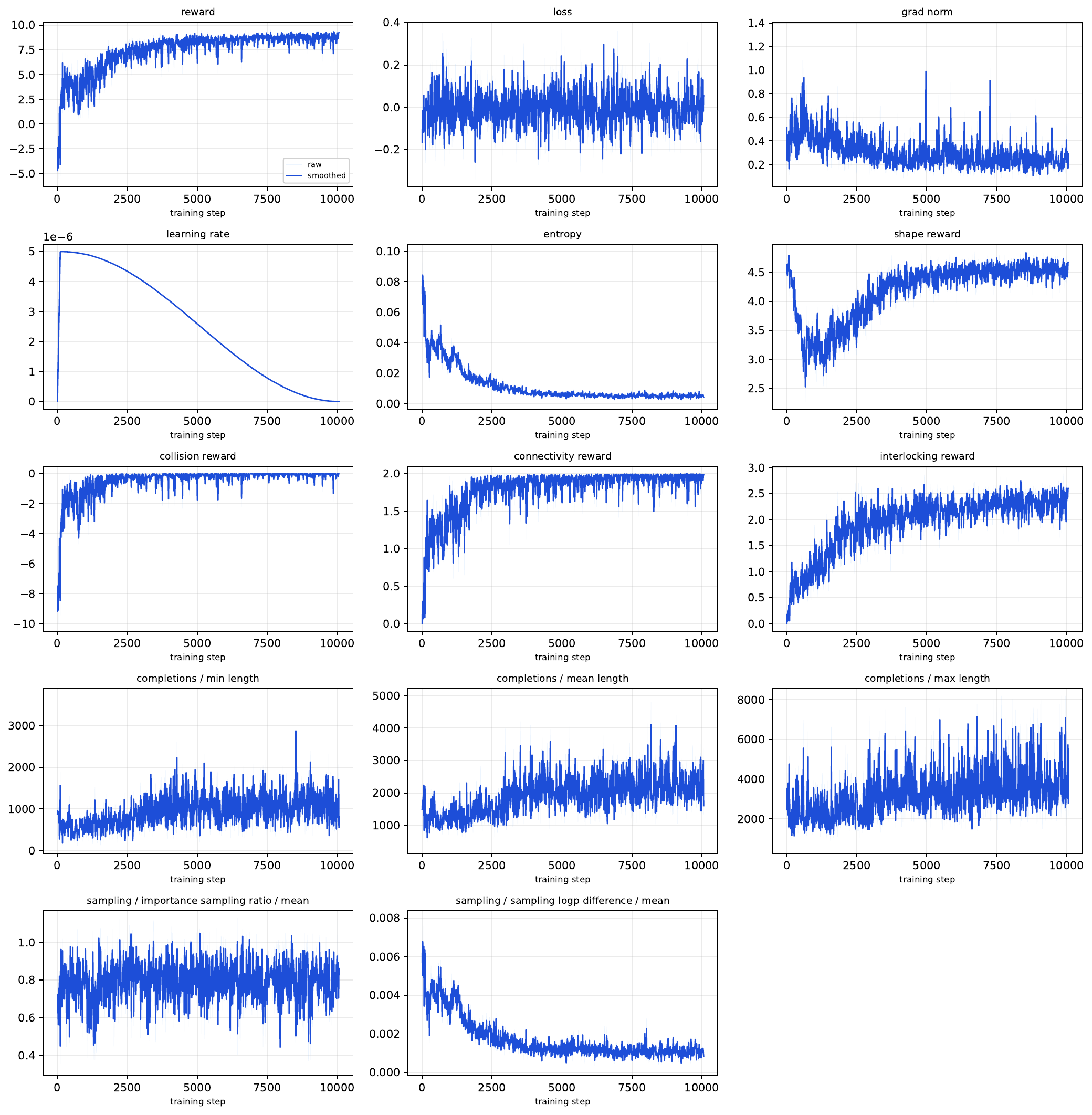}
\caption{Training dynamics of the full \texttt{STABLE} model.}
    \label{fig:dynamics_main}
\end{figure}

\begin{figure}
    \centering
    \includegraphics[width=1.0\linewidth]{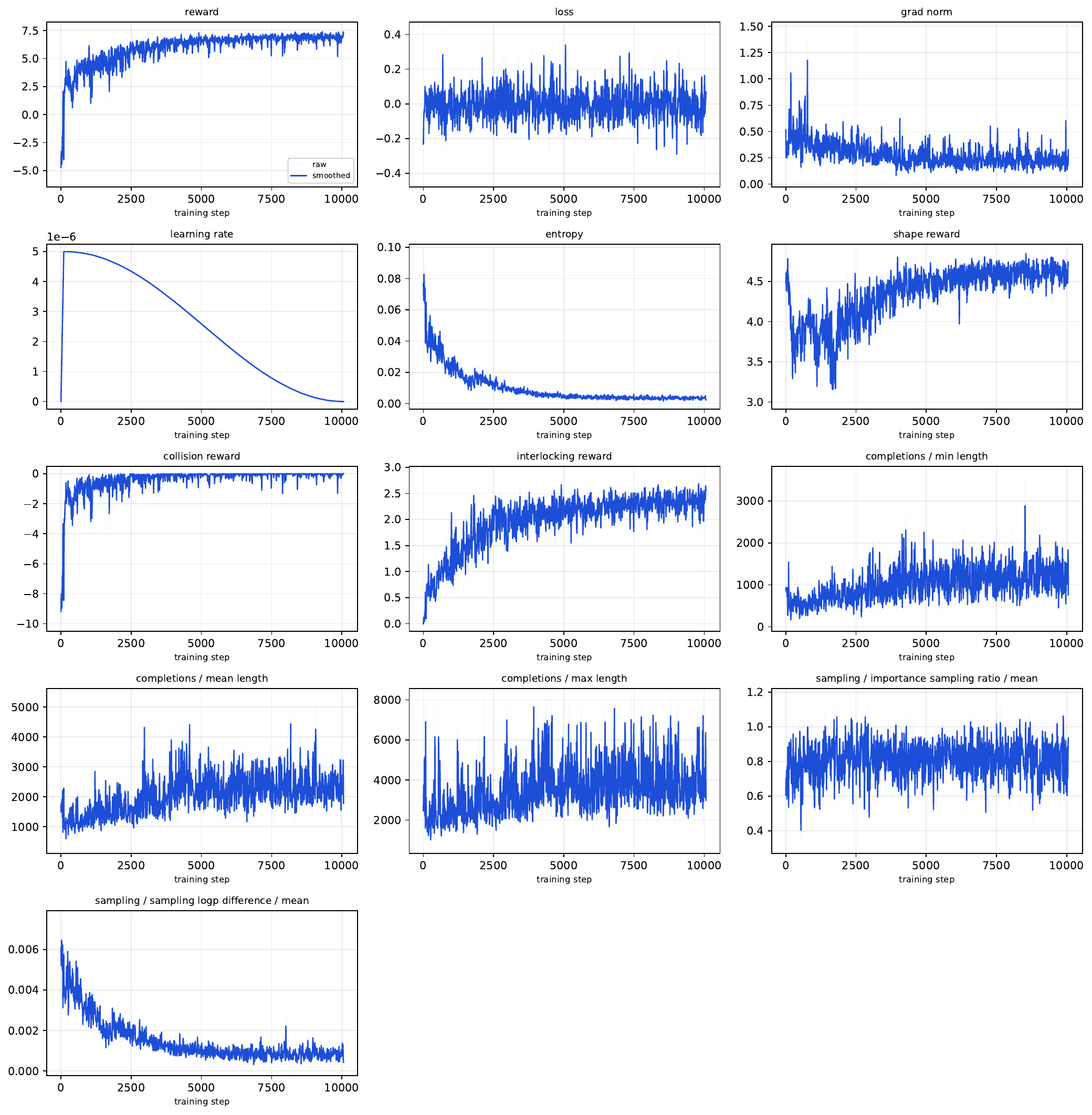}
\caption{Training dynamics of \texttt{STABLE} without the connectivity reward.}
    \label{fig:dynamics_connectivity}
\end{figure}

\begin{figure}
    \centering
    \includegraphics[width=1.0\linewidth]{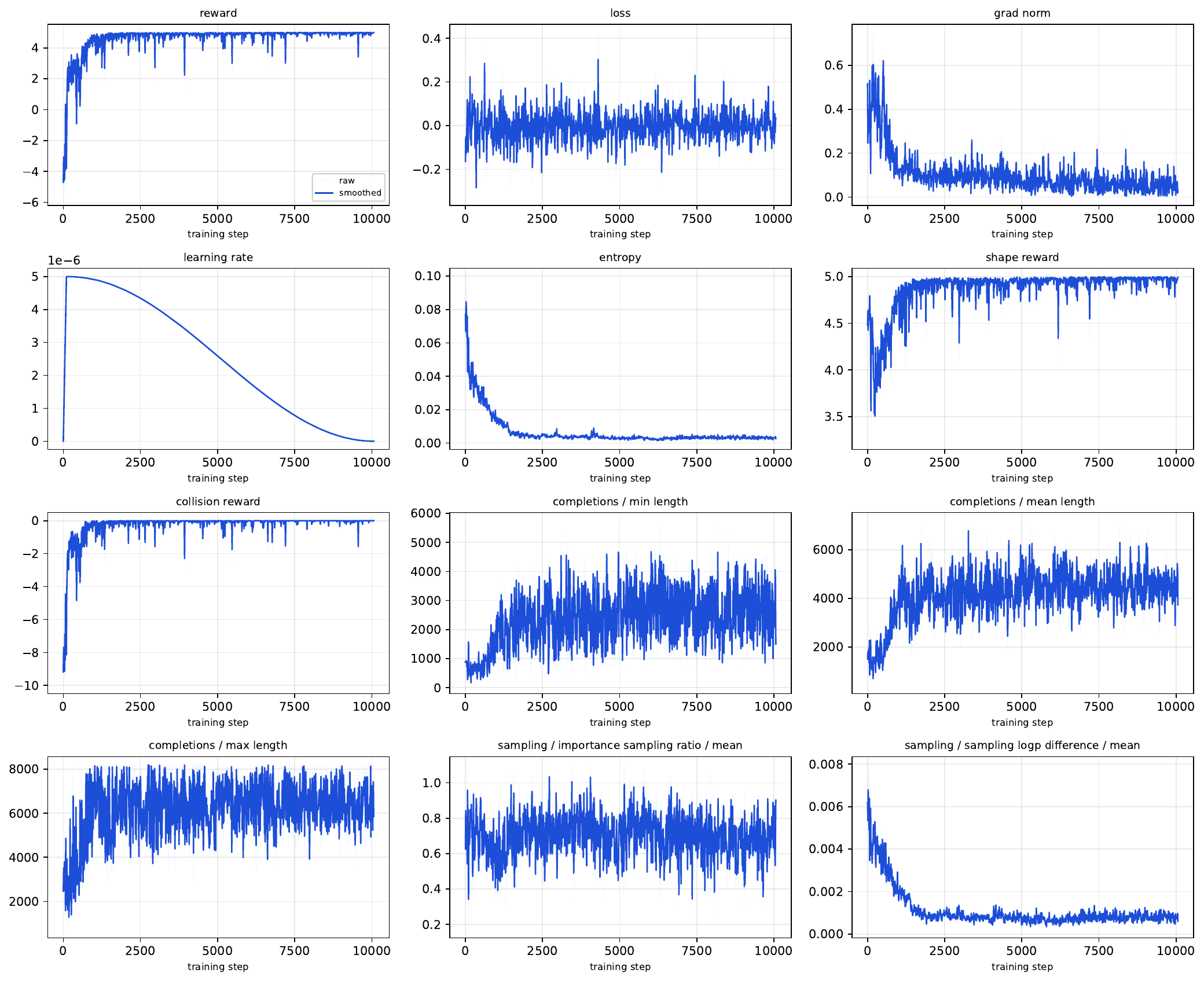}
\caption{Training dynamics of \texttt{STABLE} without structural rewards.}
    \label{fig:dynamics_structural}
\end{figure}

\begin{figure}
    \centering
    \includegraphics[width=1.0\linewidth]{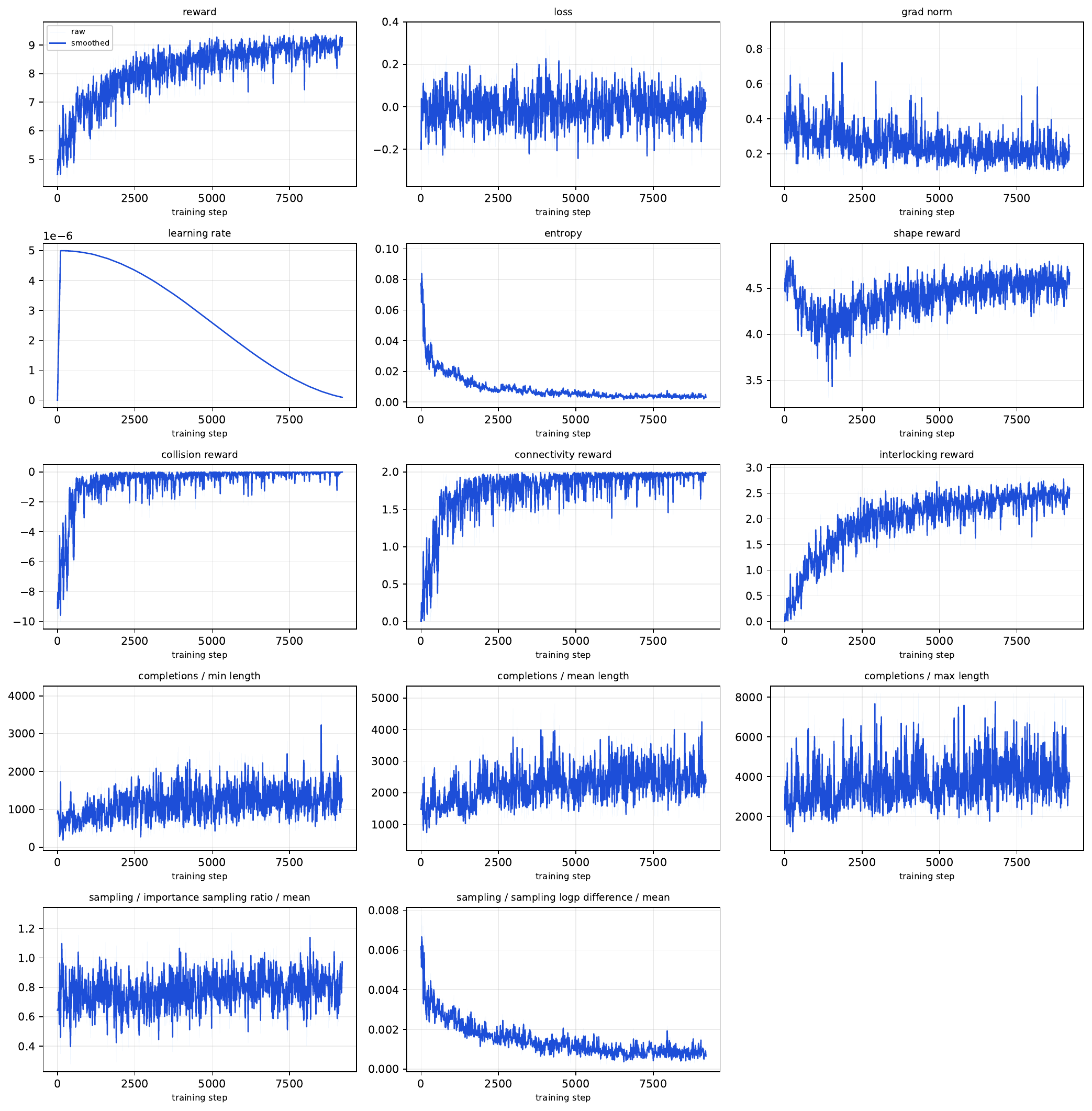}
\caption{Training dynamics of \texttt{STABLE} without the collision objective.}
    \label{fig:dynamics_collision}
\end{figure}

\begin{figure}
    \centering
    \includegraphics[width=1.0\linewidth]{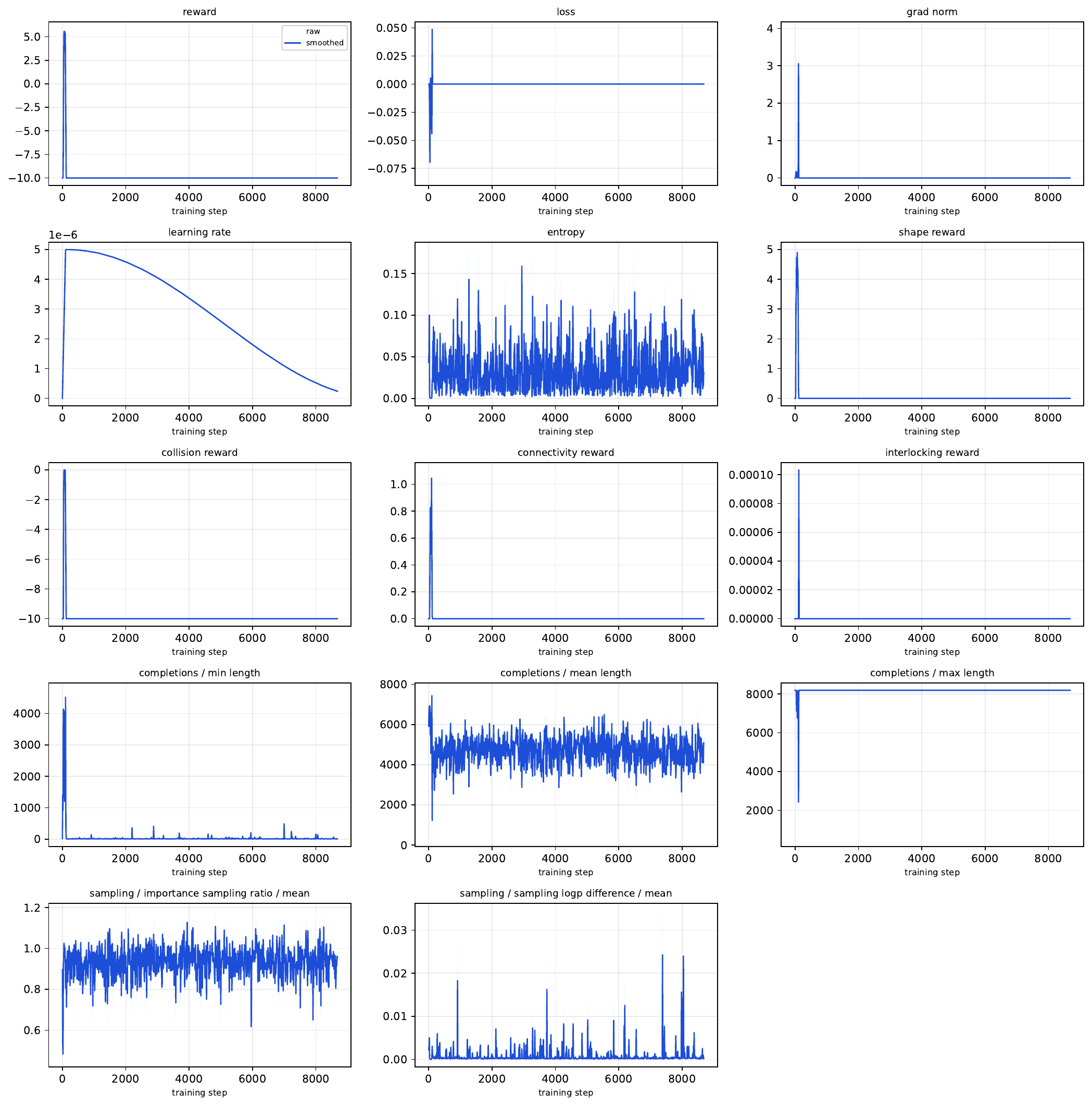}
\caption{Training dynamics of \texttt{STABLE-ZERO}.}
    \label{fig:dynamics_zero}
\end{figure}

\section{Training Dynamics}
\label{app:training_dynamics}

Fig.~\ref{fig:dynamcis_all} provides the full training dynamics of \texttt{STABLE} and its reward-ablation variants. Across all GRPO-trained models, the total reward increases during training and the entropy decreases, indicating that the policy gradually moves from exploratory generation to more stable construction behavior. The collision reward also improves rapidly in the early stage, showing that spatial feasibility is one of the first physical constraints learned by the policy.

The full \texttt{STABLE} model, shown separately in Fig.~\ref{fig:dynamics_main}, exhibits the most balanced optimization behavior. The shape reward remains high, while the connectivity and interlocking rewards steadily increase. This confirms that \texttt{STABLE} does not obtain stability by sacrificing geometric fidelity; instead, the policy learns brick placements that are simultaneously shape-consistent and structurally coherent.

The ablation dynamics further explain the quantitative results in Table~\ref{tab:reward_ablation}. Without the connectivity reward, the model still improves collision and interlocking rewards, but lacks an explicit signal for global assembly coherence. Without both structural rewards, the model mainly optimizes shape and collision, which leads to high IoU but weak connectivity and interlocking. Without the collision objective, the model can still improve shape, connectivity, and interlocking rewards, but cannot reliably eliminate voxel overlaps. These trends support the role of each reward term in constraining a different failure mode of brick generation.

As shown in Fig~\ref{fig:dynamics_zero}, the training of \texttt{STABLE-ZERO} totally fails because the base model can not produce parseable brick sequence. This means that the RL-only training can not directly induce the brick construction ability due to the sparsity of the reward. 

\section*{Broader Impact}
\label{sec:broader_impact}

This work studies how physical constraints can be internalized by generative models instead of being enforced only through test-time correction. The proposed idea may benefit applications in computer-aided design, educational construction tools, robotic assembly planning, and physically grounded 3D content generation. By reducing dependence on expensive inference-time simulation, \texttt{STABLE} may also make stable structure generation more efficient and accessible. At the same time, generated physical designs should not be directly used in safety-critical engineering or real-world load-bearing construction without expert verification. Our method is intended for research and creative assembly scenarios, and responsible deployment should include external validation when physical safety matters.

\section{Extended Related Work}
\label{app:extended_related_work}

\paragraph{Physics-aware LEGO stability analysis.}
A closely related line of work studies how to evaluate the physical stability of LEGO-like block assemblies through explicit mechanics models. These works show that buildability and stability have long been central concerns in computational brick design, but they rely on evolutionary search rather than learning a reusable generative policy. More recently, StableLego \cite{liu2024stablelego} formulates block-stacking stability analysis as an optimization problem over force-balancing equations and provides a large-scale dataset with stability inference for LEGO layouts. BrickSim \cite{wen2026bricksim} further develops a real-time physics-based simulator for interlocking brick assemblies, modeling snap-fit mechanics and structural collapse. These methods provide accurate physics-based analysis and could serve as alternative reward or evaluation oracles. In contrast, \texttt{STABLE} does not rely on simulator-based rewards during training. We instead use deterministic construction-aware rewards to expose key stability-related properties to the autoregressive policy, while evaluating final structures with both geometric and structural metrics.

\paragraph{Optimization and evolutionary LEGO design.}
Classical LEGO construction methods often formulate brick layout generation as a search or optimization problem. Legolization \cite{luo2015legolization} converts 3D models into LEGO designs by jointly considering shape approximation, color, workload, and physical stability through iterative refinement. Other works study evolutionary or genetic optimization for LEGO assemblies. Peysakhov and Regli \cite{peysakhov2003assembly} represent LEGO assemblies as labeled graphs and optimize them with genetic algorithms. These approaches explicitly search over candidate assemblies for each input and use handcrafted objectives to improve buildability. \texttt{STABLE} differs in that it learns a parametric autoregressive policy from data and reward feedback, allowing it to generate complete brick sequences directly at inference time without per-instance optimization, rollback, or evolutionary refinement.

\paragraph{Learning-based combinatorial assembly.}
Learning-based methods have recently explored LEGO-like construction as an interactive decision-making problem. Brick-by-Brick \cite{chung2021brick} formulates combinatorial construction as sequential placement of unit primitives and uses reinforcement learning with action validity prediction to satisfy geometric constraints. AssemblyComplete \cite{chen2024assemblycomplete} studies 3D combinatorial assembly completion, where an agent completes missing parts of an incomplete LEGO-like structure under collision, stability, and inventory constraints. These methods are conceptually close to \texttt{STABLE} because they also use reinforcement learning for constrained construction. However, they typically focus on interactive completion or construction policies over environment actions, often with explicit action masking or object libraries. \texttt{STABLE} instead targets point-cloud-conditioned generative modeling with a language-model backbone, where the output is a complete autoregressive brick-token sequence. Our reinforcement learning stage does not mask invalid actions at inference time, but trains the model to internalize construction constraints so that rollback-free decoding becomes possible.

\paragraph{Assembly sequence planning.}
Another related direction is assembly sequence planning, which focuses on finding feasible orders for constructing a given object. Graph Transformer based assembly planning \cite{ma2023planning} represents LEGO models as heterogeneous graphs and learns latent rules for predicting feasible and human-like assembly sequences. Such graph-based methods are complementary to our autoregressive representation. They explicitly encode part relations and assembly dependencies, whereas \texttt{STABLE} treats the brick layout itself as a sequence generated from point-cloud tokens. Incorporating graph-structured relational encodings into the policy may further improve long-range dependency modeling, but our current focus is to show that construction stability can already be internalized through reward-based post-training of an autoregressive generator.

\paragraph{Relation to \texttt{STABLE}.}
Together, these works highlight three recurring principles in brick construction. A generated assembly should be geometrically valid, globally coherent, and locally robust under inter-brick support relations. Classical optimization and simulation methods enforce these principles through explicit search or physics solvers, while interactive RL methods learn policies in constrained environments. \texttt{STABLE} combines the advantages of autoregressive generative modeling and structure-level reward optimization. It keeps the inference procedure simple and rollback-free, but uses training-time rewards to make latent construction principles explicit to the model.

\section*{Limitations}
\label{sec:limitations}

\texttt{STABLE} is currently evaluated in a fixed voxel world with a predefined brick library, which limits its coverage of larger scales, irregular components, and continuous geometries. Its rewards capture collision, connectivity, interlocking, and shape conformity, but do not model all real-world mechanics such as friction, deformation, or external forces. In addition, the method reduces inference cost by shifting computation to RL training. Extending \texttt{STABLE} to broader physical assembly domains remains future work.

\section*{LLM Usage}

LLM is used for polish the grammar and word selection of the paper. The idea, experiments and writing are completely done by the authors.

\end{document}